\pdfoutput=1
\documentclass[runningheads]{llncs}
\usepackage{makeidx}
\usepackage{graphicx}
\usepackage[caption=false]{subfig}
\usepackage{booktabs}
\usepackage{amsmath,amssymb}
\usepackage{color}
\usepackage{units}
\usepackage{cite}
\usepackage[strings]{underscore}
\usepackage[pagebackref=false,breaklinks=true,colorlinks=true,bookmarks=false,allcolors=blue]{hyperref}
\usepackage[capitalize]{cleveref}
\usepackage[misc,geometry]{ifsym}

\usepackage{fancyhdr}

\newcommand{\etal}{\textit{et al}.}
\begin{document}
	\title{Multi-view X-ray R-CNN}
	\titlerunning{Multi-view X-ray R-CNN}
	\authorrunning{J.-M. O. Steitz \etal}
	\author{Jan-Martin O. Steitz\orcidID{0000-0002-3549-312X} \and
	Faraz Saeedan\textsuperscript{\href{mailto:faraz.saeedan@visinf.tu-darmstadt.de} \Letter}\orcidID{0000-0002-8932-2983} \and
	Stefan Roth\orcidID{0000-0001-9002-9832}}
	\institute{Department of Computer Science, TU Darmstadt, Darmstadt, Germany\\
	\email{faraz.saeedan@visinf.tu-darmstadt.de}}

	\maketitle
	\thispagestyle{fancy}

	\begin{abstract}
    Motivated by the detection of prohibited objects in carry-on luggage as a part of avionic security screening, we develop a CNN-based object detection approach for multi-view X-ray image data.
    Our contributions are two-fold.
    First, we introduce a novel multi-view pooling layer to perform a 3D aggregation of 2D CNN-features extracted from each view.
    To that end, our pooling layer exploits the known geometry of the imaging system to ensure geometric consistency of the feature aggregation.
    Second, we introduce an end-to-end trainable multi-view detection pipeline based on Faster R-CNN, which derives the region proposals and performs the final classification in 3D using these aggregated multi-view features.
    Our approach shows significant accuracy gains compared to single-view detection while even being more efficient than performing single-view detection in each view.
\end{abstract}

	\section{Introduction}
\label{sec:introduction}
\begin{figure}[t!]
        \centering
        \fbox{\includegraphics[width=2.75cm,height=1.858cm]{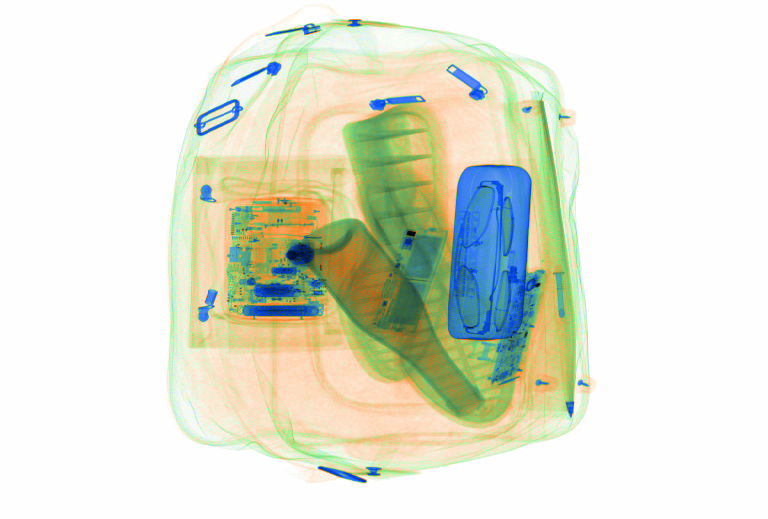}}\hspace{-\fboxrule}%
        \fbox{\includegraphics[width=2.75cm,height=1.858cm]{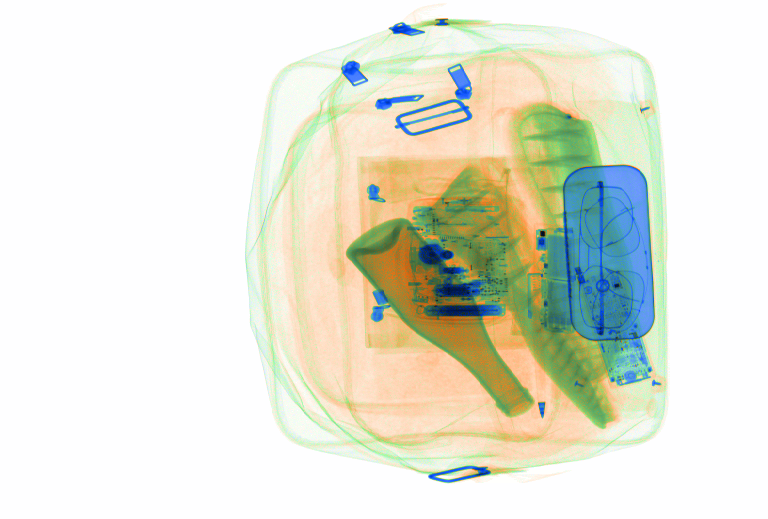}}\hspace{-\fboxrule}%
        \fbox{\includegraphics[width=2.75cm,height=1.858cm]{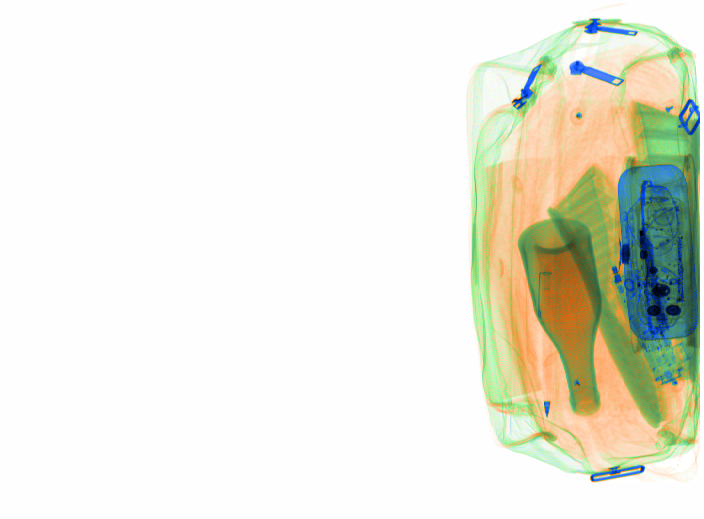}}\hspace{-\fboxrule}%
        \fbox{\includegraphics[width=2.75cm,height=1.858cm]{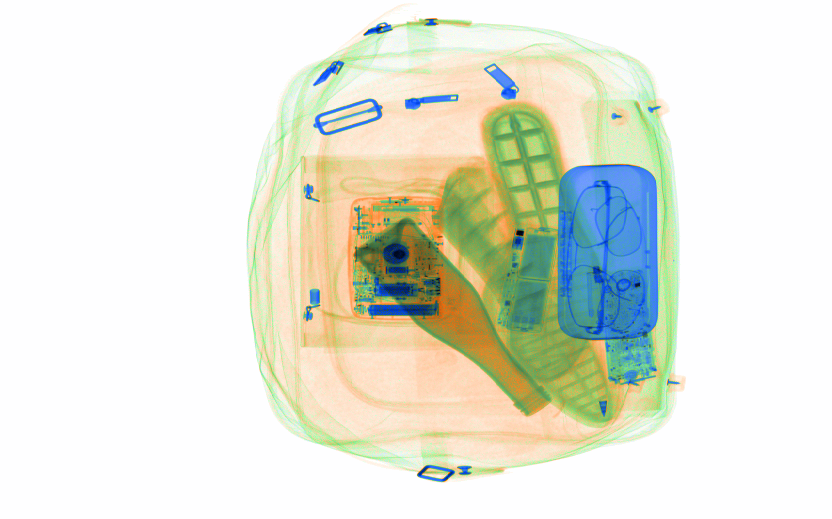}}%
        \caption{An example of multi-view X-ray images of hand luggage containing a glass bottle.}%
        \label{fig:scan}
    \end{figure}

    Baggage inspection using multi-view X-ray imaging machines is at the heart of most aviation security screening programs.
    Due to inherent shortcomings in human inspection arising from gradual fatigue, occasional erroneous judgments, and privacy concerns, computer-aided automatic detection of dangerous goods in baggage has long been a sought-after goal \cite{mery:2016:modern}.
    However, earlier approaches, mostly based on hand-engineered features and support vector machines, fell far short of providing detection accuracy comparable to human operators, which is critical due to the sensitive nature of the task \cite{Franzel2012,Bastan2015}.
    Thanks to recent advances in object detection using deep convolutional neural networks \cite{Ren2017,Girshick2015,Girshick2016,Redmon2016} with stunning success in photographic images, the accuracy of single-view object detection in X-ray images has improved significantly \cite{Jaccardicdp}.
    Yet, most X-ray machines for baggage inspection provide multiple views (two or four views) of the screening tunnel.
    An example of this multi-view data is shown in \cref{fig:scan}.
    Multi-view approaches for these applications have been only used in the context of classical methods, but whether CNN-based detectors can benefit is unclear.
    In fact, \cite{Jaccardicdp} found that a naive approach feeding features extracted from multiple views simultaneously to a fully-connected layer for detection leads to a performance drop over the single-view case.

    Fueled by applications such as autonomous driving, 3D object detection has lately gained momentum \cite{geiger:2012:kitti}.
    These 3D detection algorithms, unlike their 2D counterparts, are not general purpose and rely on certain sensor combinations or employ heavy prior assumptions that make them not directly applicable to multi-view object detection in X-ray images.
    Some of these 3D detection algorithms assume that the shape of the desired object is known in the form of a 3D model \cite{aubry:2014:seeing3d,romea:2011:moped}.
    Yet, 3D models of objects are more difficult to acquire compared to simple bounding box annotations and a detector that relies on them for training may not generalize well on objects with highly variable shape such as handguns.
    Other methods use point clouds from laser range finders alone or in conjunction with RGB data from a camera \cite{qi:2018:frustum,chen:2017:mv3d}.
    Our setup, in contrast, provides multi-view images of objects in a two channel (dual-energy) format, which is rather different from stereo or point cloud representations.
    Using 3D convolutions and directly extending existing 2D methods is one possibility, but the computational cost and memory requirements can be prohibitive when relying on very deep CNN backbone architectures such as ResNet \cite{resnet}.

    In this work, we extend the well-known Faster R-CNN \cite{Ren2017} to multi-view X-ray images by employing the idea of late fusion, which enables our use of very deep CNNs to extract powerful features from each individual view, while deriving region proposals and performing the classification in 3D.
    We introduce a novel \emph{multi-view pooling layer} to enable this fusion of features from single views using the geometry of the imaging setup.
    This geometry is assumed fixed all through training and testing and needs to be calculated once.
    We do not assume further knowledge of the detected objects as long as we have sufficient bounding box annotations in 2D.
    We show that our method, termed \emph{MX-RCNN}, is not only highly flexible in detecting various hazardous object categories with very little extra knowledge required, but is also considerably more accurate while even being more efficient than performing single-view detection in the four views.

	\section{Related Work}
    Inspired by the impressive results in 2D object detection, several recent works \cite{tulsiani:2015:viewpoint,xiang:2015:3dvp,Xiang:2017:subcnn,mousavian:2016:3ddeepbox} build upon a 2D detection in an image and then attempt to estimate the 3D object pose.
    These methods, however, rely predominantly on a set of prior constraints such as objects being located on the ground for estimating the 3D position of the object.
    These prior constraints do not appear easily extensible to our problem case, since objects inside bags can take an arbitrary orientation and position.

    Other approaches have tried to work directly with depth data \cite{song:2014:sliding,song:2016:deepsliding,li:2016:3dfcn}, where most methods voxelize the space into a regular grid and apply 3D convolutions to the input.
    While this yields the most straightforward extension of well-proven 2D detectors, which are based on 2D convolution layers, increasing the dimensionality of convolution layers can only be done at very low resolution as denser voxelizations of the space result in unacceptably large memory consumption and processing time.
    Our method also uses the idea of 3D convolutions but defers them to very late stages in which we switch from 2D to 3D.
    In doing so, we enable the detector to leverage high image resolution in the input while extracting powerful features that leverage view-consistency.

    A number of methods use the geometry of the objects as prior knowledge to infer a 6-DoF object pose.
    These methods mainly rely on CAD models or ground truth 3D object models and match either keypoints between these models and 2D images \cite{aubry:2014:seeing3d,romea:2011:moped,zhu:2014:grasping} or entire reconstructed objects \cite{rothganger:2006:affine}.
    Borrowing ideas from robotics mapping, the estimated pose of an object produced by a CNN can also be aligned to an existing 3D model using the iterative closest points (ICP) algorithm \cite{gupta:2015:align}.

    Hang \etal \cite{hang:2015:shape} proposed a method most closely related to ours, which aggregates features extracted from multiple views of a scene using an elementwise max operation among all views.
    Yet, unlike our multi-view pooling layer this aggregation is not geometry-aware.

    Despite the recent focus on the problem of visual object detection, its application to X-ray images has not received as much attention.
    Some older methods \cite{Franzel2012,Bastan2015} exist for this application, but they perform considerably weaker than deep learning-based methods \cite{Akcay:2016:tlxray}.
    However, the use of CNNs on X-ray images for baggage inspection has been limited to the direct application of basic 2D detection algorithms, either with pretraining on photographic images or training from scratch.
    Jaccard \etal \cite{Jaccardicdp} propose a black-box approach to multi-view detection by extracting CNN features from all views, concatenating them, and feeding them to fully-connected layers.
    Yet, the accuracy fell short of that of the original single-view detection.
    To the best of our knowledge, there exists no previous end-to-end learning approach that successfully uses the geometry of the X-ray machine to perform a fusion of features from various views.

	\section{Multi-view X-ray R-CNN}
\label{sec:theory}
	We build on the standard Faster R-CNN object detection method \cite{Ren2017}, which works on single-view 2D images and is composed of two stages.
	They share a common feature extractor, which outputs a feature map.
	The first stage consists of a Region Proposal Network (RPN) that proposes regions of interest to the second stage.
	Those regions are then cut from the feature map and individually classified.
	The RPN uses a fixed set of 9 standard axis-aligned bounding boxes in 3 different aspect ratios and 3 scales, so-called anchor boxes.
	Those anchor boxes are defined at every location of the feature map.
	The RPN then alters their position and shape by learning regression parameters in training.
	Additionally, a score to distinguish between objects and background is learned in a class-agnostic way, which is used for non-maximum suppression and to pass only a subset of top-scoring region proposals to the second stage.
	The second stage classifies the proposals and outputs additional regression values to fine-tune the bounding boxes at the end of the detection process.

	\subsubsection{MX-RCNN.} The basic concept of our Multi-view X-ray R-CNN (MX-RCNN) approach is to perform feature extraction on 2D images to be able to utilize standard CNN backbones including ImageNet \cite{Russakovsky2015} pretraining.
	This addresses the fact that the amount of annotated X-ray data is significantly lower than that of photographic images.
	We then combine the extracted feature maps of different projections, or views, into a common 3D feature space in which object detection takes place, identifying 3D locations of the detected objects.

	\begin{figure}[t]
	\centering
		\includegraphics[width=\linewidth]{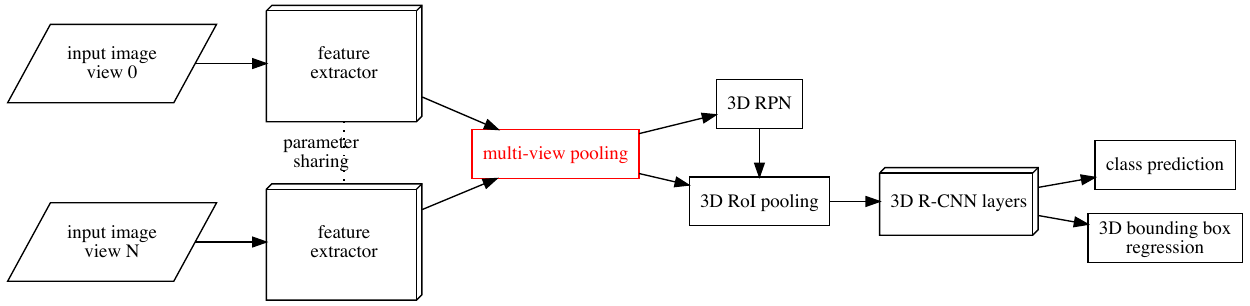}
	    \caption{\emph{Schema of the hybrid 2D-3D MX-RCNN architecture.} The features of each view are extracted independently in 2D and combined in the multi-view pooling layer (\emph{marked in red}). The resulting common 3D feature volume is passed to the RPN and to the RoI pooling layer where regions are extracted for evaluation in the 3D R-CNN layers.}
		\label{fig:faster-rcnn}
	\end{figure}

	Our MX-RCNN uses a ResNet-50 \cite{He2016} architecture, where the first 4 out of its 5 stages are used for feature extraction on the 2D images.
 	Then a novel multi-view pooling layer, provided with the fixed geometry of the imaging setup, combines the feature maps into a common 3D feature volume.

	The combined feature space is passed to a RPN, which has a structure similar to the RPN in Faster R-CNN \cite{Ren2017}, but with 3D convolutional layers instead of 2D ones.
	Further, it has \(6 \times A\) regression parameter outputs per feature volume position, where \(A\) is the number of anchor boxes, because for 3D bounding boxes 6 regression parameters are needed.
	Following Faster R-CNN, we define those regression parameters as
	\begin{equation}
		\begin{aligned}
			t_x &= \nicefrac{(x-x_a)}{w_a} \quad& t_y &= \nicefrac{(y-y_a)}{h_a} \quad& t_z &= \nicefrac{(z-z_a)}{d_a} \\[.5em]
			t_w &= \log(\nicefrac{w}{w_a}) \quad& t_h &= \log(\nicefrac{h}{h_a}) \quad& t_d &= \log(\nicefrac{d}{d_a}) \;,
		\end{aligned}
	\end{equation}
	where the index \(a\) denotes the parameters of the anchor box and with bounding box center (\(x\), \(y\), \(z\)), width \(w\), height \(h\), and depth \(d\).

	The RPN proposes volumes to be extracted by a Region-of-Interest pooling layer with an output of size 7\(\times\)7\(\times\)7 to cover a part of the feature volume similar in relative size to the 2D case.
	The 3D regions are then fed into a network similar to the last stage of ResNet-50 in which all convolutional and pooling layers are converted from 2D to 3D and the size of the last pooling kernel is adjusted to fit the feature volume size.
	In contrast to the 2D stages, these 3D convolutions are trained from scratch since ImageNet pretraining is not possible.
	Afterwards, the regions are classified and 3D bounding box regression parameters are determined.
	A schema of our MX-RCNN is depicted in \cref{fig:faster-rcnn}.

\subsection{K-means clustering of anchor boxes}
	When we expand the hand-selected aspect ratios of the Faster R-CNN anchor boxes of 1:1, 1:2, and 2:1 at 3 different scales to 3D, we arrive at a total of 21 anchor boxes.
	Since this number is large and limits the computational efficiency, we aim to improve over these standard anchor boxes.
	To that end, we assess the quality of the anchor boxes as priors for the RPN.
	Specifically, we use their intersection over union (IoU) \cite{Jaccard1912} with the ground-truth annotations of the training set.
	We instantiate anchor boxes at each position of the feature map used by the RPN and for each ground-truth annotation we find its highest IoU with an anchor box.
	For the standard anchor boxes expanded to 3D, this yields an average IoU of 0.5.

\begin{figure}[t]
	\centering
		\includegraphics[height=4.5cm]{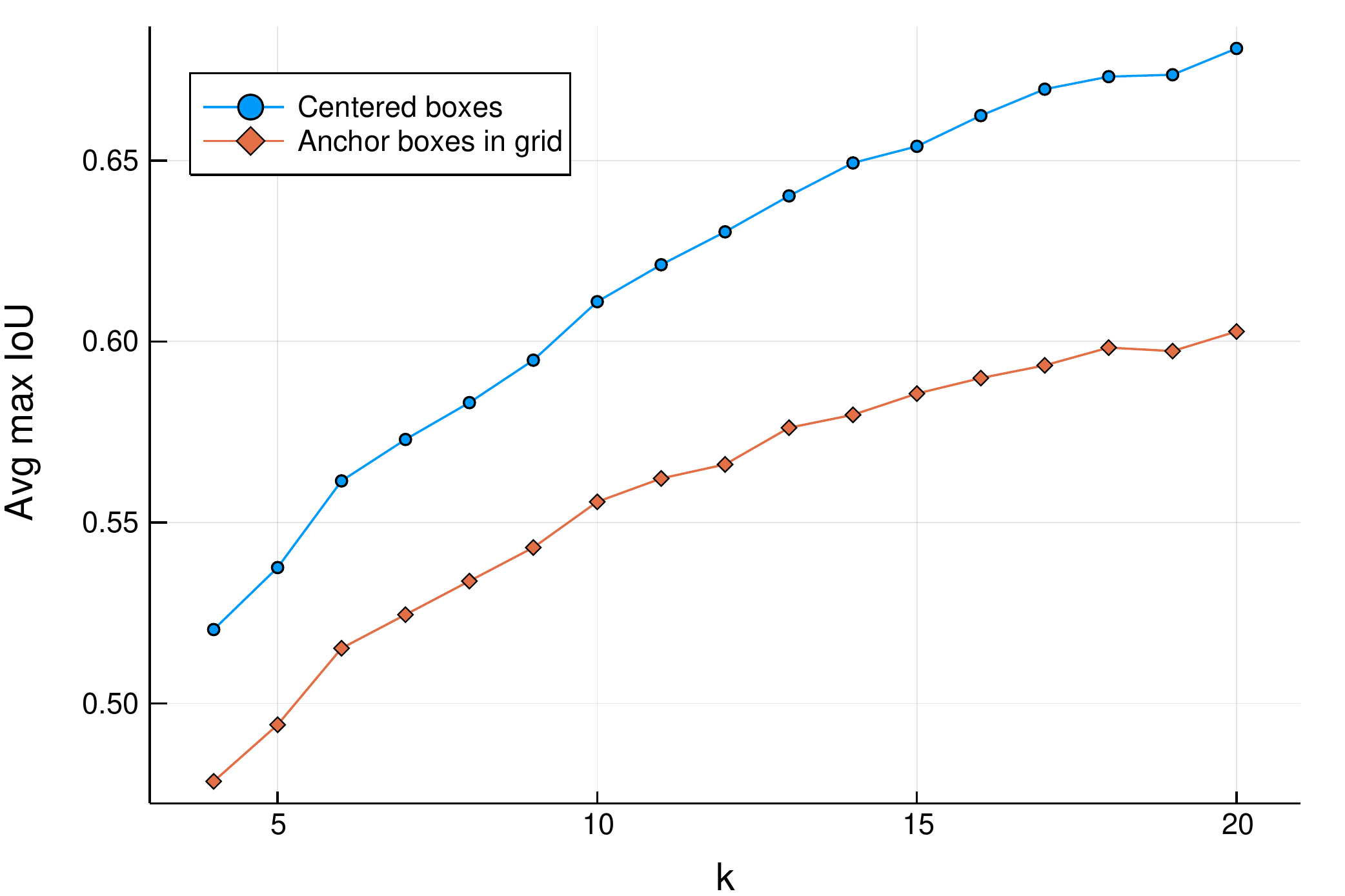}
	    \caption{\emph{Average IoU between ground-truth boxes and the closest cluster from $k$-means clustering as a function of $k$.} The IoU is calculated for anchor boxes centered on the ground-truth boxes (\textit{blue, circles}) and for anchor boxes distributed in the grid of the feature map (\textit{red, diamonds}). Even a low number of clusters already outperforms the standard hand-selected anchor boxes (IoU of 0.5).}
		\label{fig:k-means}
	\end{figure}

	To improve upon this while optimizing the computational efficiency, we follow the approach of the YOLO9000 object detector \cite{Redmon2016} and use $k$-means clustering on the bounding box dimensions (width, height, depth) in the training set to find priors with a better average overlap with the ground-truth annotations.
	We employ the Jaccard distance \cite{Jaccard1912}
	\begin{equation}
	    d_\mathrm{J}(a, b) = 1 - \mathrm{IoU}(a, b)
	\end{equation}
	between boxes \(a\) and \(b\) as a distance metric.
	We run $k$-means clustering for various values of \(k\).
	\cref{fig:k-means} shows the average IoU between ground-truth bounding boxes and the closest cluster (\textit{blue, circles}).
	To convert clusters into anchor boxes, they have to be aligned to the resolution of a feature grid.
	To account for this, we also plot the IoU between the ground-truth bounding boxes and the closest cluster once it has been shifted to the nearest feature grid position (\textit{red, diamonds}).
	We choose \(k=10\), which achieves an average IoU of 0.56 for the resulting anchor boxes distributed in a grid.
	This is clearly higher than using the standard 21 hand-selected anchor boxes, while maintaining the training and inference speed of a network with only 10 anchor boxes.

\subsection{Multi-view pooling}
\begin{figure}[t]
	\centering
		\includegraphics[height=6cm]{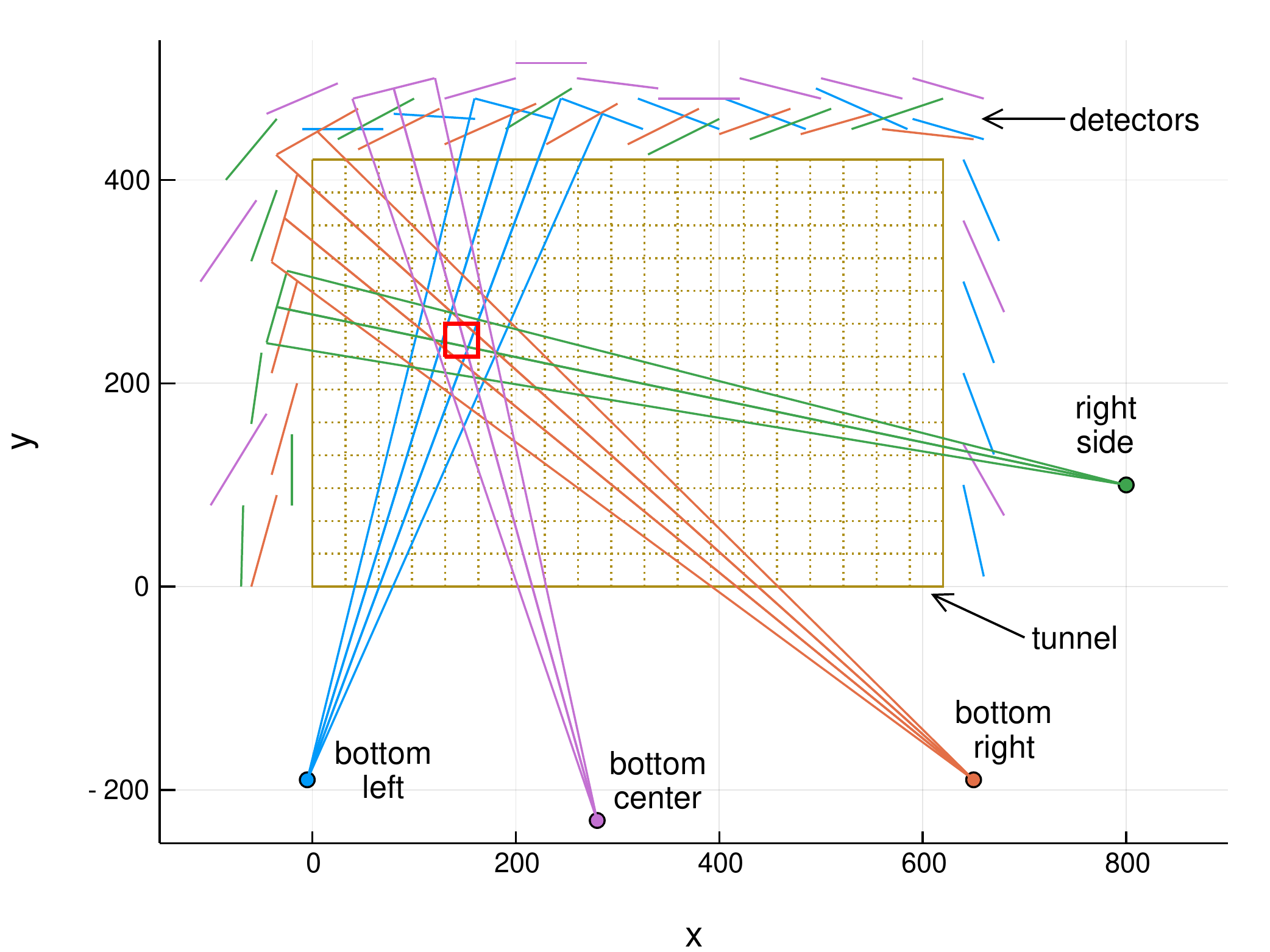}
	    \caption{Example plot illustrating the relevant beams (\textit{color-coded per view}) in the multi-view pooling of a specific output cell (\textit{marked in red}). The actual geometry differs slightly.}
		\label{fig:mv-pooling}
	\end{figure}

	Our proposed multi-view pooling layer maps the related feature maps of the 2D views of an X-ray recording into a common 3D feature volume by providing it with the known geometry of the X-ray image formation process in the form of a weight matrix.
	To determine the weights, we connect each group of detector locations related to one pixel in the 2D feature map to their X-ray source to form beams across the 3D space.
	For each of the output cells of the 3D feature volume, we use the volume of their intersection with the beams normalized by the cell volume as relative weight factors.
	The multi-view pooling layer then computes the weighted average of the feature vectors of all beams for each output cell, normalized by the number of views in each X-ray recording; we call this variant MX-RCNN\textsubscript{avg}.
	Additionally, we implemented a version of the multi-view pooling layer that takes the maximum across the weighted feature vectors of all beams for each output cell; we call this variant MX-RCNN\textsubscript{max}.
	An example of a mapping, specific to our geometry, is shown in \cref{fig:mv-pooling}.

\subsection{Conversion of IoU thresholds}
\label{subsec:iou-conversion}
	Since the IoU in 3D (volume) behaves differently than in 2D (area), we aim to equalize for this.
	Specifically, we aim to apply the same strictness for spatial shifts that are allowed per bounding box dimension such that a proposed object is still considered a valid detection.
	We assume that the prediction errors of the bounding box regression values are equally distributed across all dimensions.
	For simplicity, we further assume that errors are only made up of shifts compared to the ground-truth bounding boxes.
	In 2D, the allowed relative shift \(s\) per dimension of a bounding box of arbitrary dimensions for an IoU threshold of \(t_2\) is given by
	\begin{equation}
		s(t_2) = 1 - \sqrt{\nicefrac{2 t_2}{(t_2 + 1)}} \;.
	\end{equation}
	The same relative shift applied to an arbitrarily sized 3D bounding box would require the threshold
	\begin{equation}
		t_3(s) = \nicefrac{(1-s)^3}{(2-(1-s)^3)} \;.
	\end{equation}
	An IoU threshold \(t_2\) applied to the 2D case therefore becomes
	\begin{equation}
		t_3(t_2) = \frac{\sqrt{\nicefrac{2 t_2}{(1+t_2)}}^3}{2 - \sqrt{\nicefrac{2 t_2}{(1+t_2)}}^3}
	\end{equation}
	for use with 3D bounding boxes if the same strictness per dimension is to be maintained.
	The evaluation of the \textsc{Pascal} VOC challenge \cite{Everingham2010} for detection in 2D images uses a standard threshold of 0.5, which yields a threshold of 0.374 in the 3D detection case.

\subsection{Computational cost}
	Our single-view Faster R-CNN implementation reaches a frame rate of \unit[3.9]{fps} for training and \unit[6.1]{fps} for inference on a NVIDIA GeForce GTX Titan X GPU.
	4 frames or images need to be processed for one complete X-ray recording.
	Our MX-RCNN achieves frame rates of \unit[4.6]{fps} for training and \unit[7.6]{fps} for inference; note that it can share work across the 4 frames as they are processed simultaneously.
	Thus, our method is 18\% faster in training and 25\% faster in inference.
	We attribute this to the lower number of regions extracted per recording, because of the common classification stage in the multi-view detector, despite the higher computational costs for its 3D convolutional layers.

	\section{Multi-view X-Ray Dataset}
\label{sec:dataset}
\begin{figure}[t]
        \centering
        \subfloat[Bottom left view]{
            \fbox{\includegraphics[height=3cm]{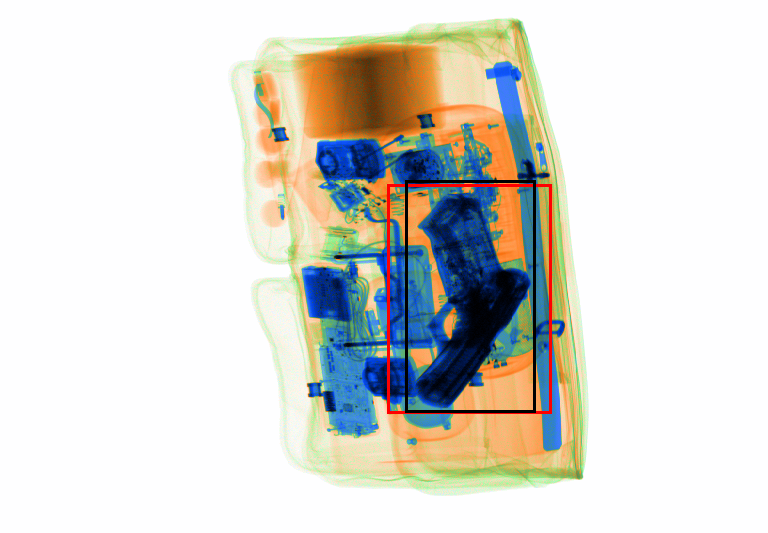}}
        } \qquad
        \subfloat[Bottom right view]{
            \fbox{\includegraphics[height=3cm]{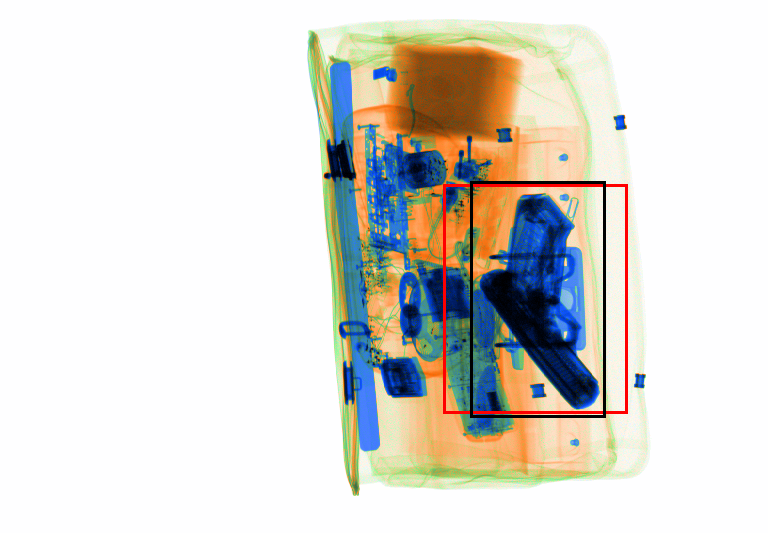}}
        } \\
        \subfloat[Right side view]{
            \fbox{\includegraphics[height=3cm]{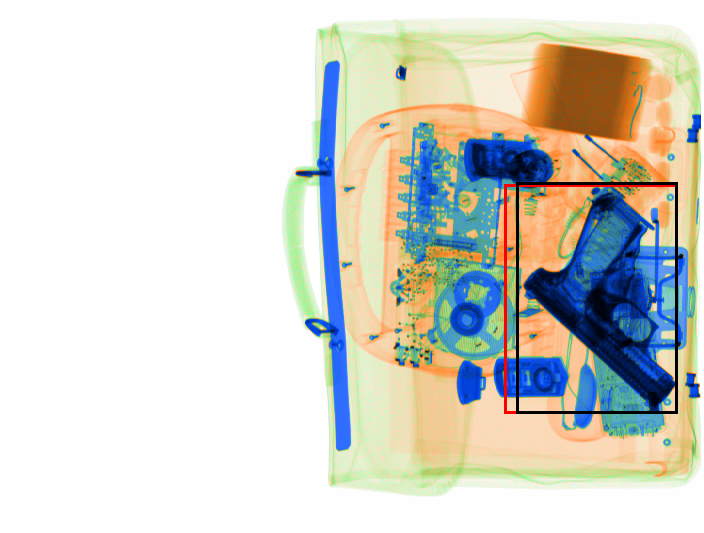}}
        } \qquad
        \subfloat[Bottom center view]{
            \fbox{\includegraphics[height=3cm]{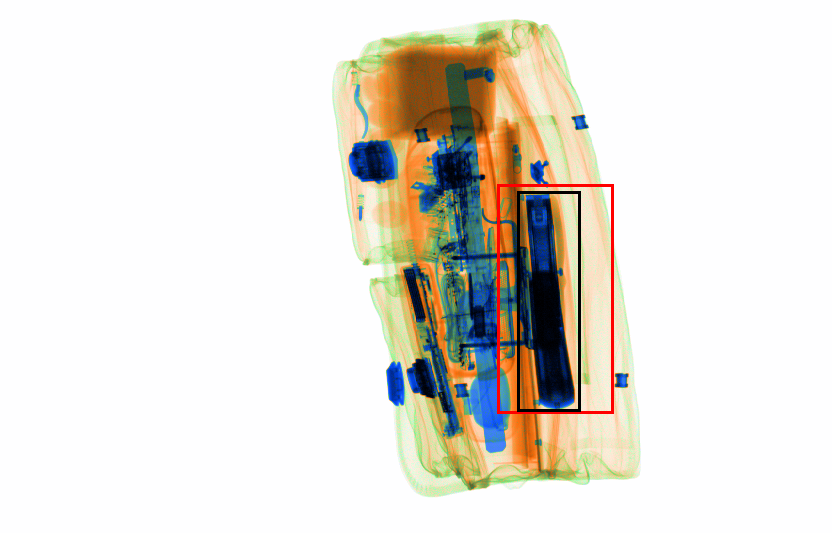}}
        }
        \caption{\emph{False-color images of all views of an example X-ray recording of baggage containing a handgun.} The handgun is easier to spot in certain images depending on their angle of projection. Bounding boxes show the original 2D annotations (\textit{black}) and the reprojected 3D annotations (\textit{red}).}
        \label{fig:example-imgs}
    \end{figure}

    Lacking a standardized public dataset for this task, we leverage a custom dataset\footnote{Unfortunately, we are not able to release the dataset to the public. Researchers wishing to evaluate on our dataset for comparison purposes are invited to contact the corresponding author.} of dual-energy X-ray recordings of hand luggage made by an X-ray scanner used for security checkpoints.
    The X-ray scanner uses line detectors located around the tunnel through which the baggage passes.
    Its pixels constitute the $x$-axis in the produced image data while the movement of the baggage, respectively the duration of the X-ray scan and the belt speed, define the $y$-axis of the image.
    Each recording consists of four different views, three from below and one from the side of the tunnel the baggage is moved through.
    The scans from each view produce two grayscale attenuation images from the dual-energy system, which are converted into a false-color RGB image that is used for the dataset.
    This is done to create images with 3 color channels that fit available models pretrained on ImageNet-1000 \cite{Russakovsky2015}.
    An example recording is shown in \cref{fig:example-imgs}.

    The following types of recordings are available in the dataset: \\
    \textbf{Glass Bottle.} Recordings of baggage containing glass bottles of different shapes and sizes. \\
    \textbf{TIP Weapon.}
    Synthetic recordings of baggage where a pre-recorded scan of a handgun is randomly projected onto a baggage recording by a method called Threat Image Projection (TIP) \cite{TIPSmiths}. A limited set of handguns is repeatedly used to generate all recordings. \\
    \textbf{Real Weapon.} Recordings that contain a handgun of various types and are obtained using a conventional scan without the use of TIP. \\
    \textbf{Negative.} Recordings containing neither handguns nor glass bottles.

    The synthetic TIP images are only used for training and validation; the complete scans with real weapons are used for testing to evaluate if the trained network generalizes.
    A detailed overview of the different subsets of the data is given in \cref{tbl:dataset}.
    The dataset is split into its subsets such that views belonging to one and the same recording are not distributed over different subsets.
    \textsc{Pascal} VOC-style annotations \cite{Everingham2010} with axis-aligned 2D bounding boxes per view are available for two different classes of hazardous objects, \textit{weapon} and \textit{glassbottle}.

    \begin{table}[t]
        \caption{Number of recordings (images) in the different subsets of the dataset.}
        \label{tbl:dataset}
        \centering
        \newcommand{\colsep}{\hspace{.5\tabcolsep}}
        \begin{tabular}{@{}l@{\hspace{1cm}}r@{\colsep}lr@{\colsep}lr@{\colsep}lr@{\colsep}l@{}}
            \toprule
            Type / Subset & \multicolumn{2}{c}{Train} & \multicolumn{2}{c}{Validation} & \multicolumn{2}{c}{Test} & \multicolumn{2}{c}{Total} \\
            \midrule
            Glass Bottle  &   358 & (1432)  &   40 & (160)   &   209 & (836)   &   607 & (2428) \\
            TIP Weapon    &  1944 & (7776)  &  216 & (864)   &     0 & (0)     &  2160 & (8640) \\
            Real Weapon   &     0 & (0)     &    0 & (0)     &   464 & (1856)  &   464 & (1856) \\
            Negative      &     0 & (0)     &    0 & (0)     &   950 & (3800)  &   950 & (3800) \\
            \midrule
            Total         &  2302 & (9208)  &  256 & (1024)  &  1623 & (6492)  &  4181 & (16724) \\
            \bottomrule
        \end{tabular}
    \end{table}

\subsection{3D bounding box annotations}
\label{subsec:generation}
    To be able to train and evaluate our multi-view object detection with 3D annotations, we generate those out of more commonly available 2D bounding box annotations.
	Specifically, we generate axis-aligned 3D bounding boxes from several axis-aligned 2D bounding boxes.
    Because all our X-ray recordings have at most one annotated object,\footnote{
    The number of annotated objects is a restriction of the dataset only;
    our detector is able to handle multiple objects per image.}
    there is no need to match multiple annotations across the different views.
    In case of multiple annotated objects per image, a geometrically consistent matching could be used.

\begin{figure}[t]
	\centering
		\includegraphics[height=6cm]{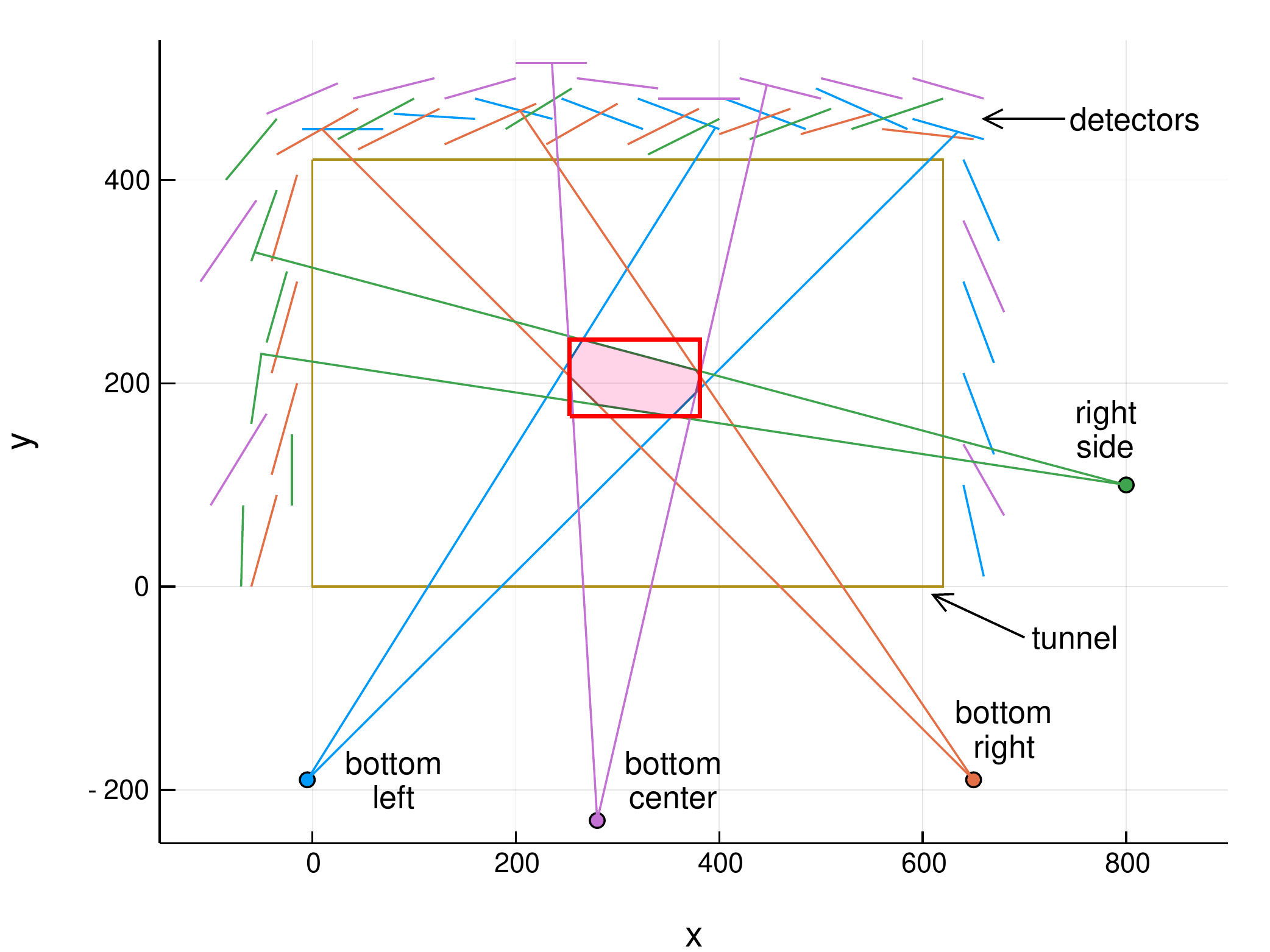}
	    \caption{\emph{Example of the 3D bounding box generation.} The lines of projection (\textit{color-coded per view}) for the 2D bounding box positions overlap in the $xy$-plane of the 3D volume. The resulting polygon of their intersection (\textit{pink}) is enclosed by the 3D bounding box (\textit{red}). The actual geometry differs slightly.}
		\label{fig:3D-bboxes}
	\end{figure}

    Recall the specific imaging setup from above.
    If we now align the 2D $y$-axis (belt direction) to the 3D $z$-axis, the problem of identifying a suitable 3D bounding box reduces to a mapping from the 2D $x$-axis to a $xy$-plane in 3D.
	The lines of projection between the X-ray sources and the detectors corresponding to the $x$-axis limits of the 2D bounding boxes define the areas where the object could be located in the $xy$-plane per view.
	We intersect those triangular areas using the Vatti polygon clipping algorithm \cite{Vatti1992} and choose the minimum axis-aligned bounding box containing the resulting polygon as an estimation of the object's position in 3D space.
	For the $z$-axis limits of the 3D bounding box, we take the mean of the $y$-limits of the 2D bounding boxes.
	An example of the generation process is shown in \cref{fig:3D-bboxes}.
	Note that while our process of deriving 3D bounding boxes is customized to the baggage screening scenario, analogous procedures can be defined for more general imaging setups.

	Note that in the ideal case, all projection lines would intersect in the 4 corners of the bounding box.
	However, due to variances in the annotations made independently for each view, this does not hold in practice.
	As a result, the bounding box enclosing the intersection polygon is an upper bound of the estimated position of the object in 3D space.
	We thus additionally project the 3D bounding boxes back to the 2D views to yield 2D bounding boxes that include the geometry approximation made in the generation of the 3D bounding box.
	The difference between an original 2D annotation and a reprojected 3D bounding box can be seen in \cref{fig:example-imgs}.
	We use these 2D annotations to train a single-view Faster R-CNN as a baseline that assumes the same object localization as the multi-view networks trained on the 3D annotations.
	If more precise 3D bounding boxes are desired, they could be obtained from joint CT and X-ray recordings, which are becoming more common in modern screening machines.

\section{Experiments}
\label{sec:experiments}

\subsection{Training}
    We scale all images used for training and evaluation by a factor of 0.5 so they have widths of 384, 384, 352, and \unit[416]{px} for the views 0 through 3 of each X-ray recording.
    For all experiments we use a ResNet-50 \cite{resnet} with parameters pretrained on ImageNet-1000 \cite{Russakovsky2015}.
    The first two stages of the ResNet-50 are fixed and we fine-tune the rest of the parameters.
    If not mentioned otherwise, hyper-parameters remain unchanged against the standard implementation of Faster R-CNN.
    All networks are trained by backpropagation using stochastic gradient descent (SGD).

    As a baseline we train a standard implementation of single-view Faster R-CNN on the training set without data augmentation.
    Single-view Faster R-CNN detects objects on all views independently.
    We start with a learning rate of 0.001 for the first 12 epochs and continue with a rate of 0.0001 for another 3 epochs.

    For training and evaluation of our MX-RCNN, we use a mini-batch size of 4 with all related views inside one mini-batch.
    We reduce the number of randomly sampled anchors in each 3D feature volume from 256 to 128 to better match the desired ratio of up to 1:1 between sampled positive and negative anchors.
    Additionally, we reduce the number of sampled RoIs to 64 to better match the desired fraction of 25\% foreground regions that have an IoU of at least 0.5 with a 3D ground-truth bounding box.
    We accordingly reduce the learning rate by the same factor \cite{Goyal2017}.
    In the multi-task loss of the RPN, we change the balancing factor \(\lambda\) for the regression loss and the normalization by the number of anchor boxes to an empirically determined factor of \(\lambda = 0.05\) without additional normalization.
    In practice, this ensures good convergence of the regression loss.
    We initialize all 3D convolutional layers as in the standard Faster R-CNN implementation.
    We then train MX-RCNN\textsubscript{avg} on the training set for 28 epochs with a learning rate of 0.0005.
    A cut of the learning rate did not show any benefit when evaluating with the validation set.
    Additionally, we train MX-RCNN\textsubscript{max} for 17 epochs with the same learning rate.
    Again, a learning rate cut did not show any further improvement on the validation set.

\subsection{Evaluation criteria}
    Since there is no established evaluation criterion for our experimental setting, we are using the average precision (AP), which is the de-facto standard for the evaluation of object detection tasks.
    Specifically, we compare the trained networks with the evaluation procedure for object detection of the \textsc{Pascal} VOC challenge that is in use since 2010 \cite{Everingham2014}.
    To be considered a valid detection, a proposed bounding box must exceed a certain threshold with a ground-truth bounding box.
    If multiple proposed bounding boxes match the same ground-truth bounding box, only the one with the highest confidence is counted as a valid detection.
    The \textsc{Pascal} VOC challenge uses an IoU of 0.5 as threshold for the case of object detection in 2D images.
    As discussed in \cref{subsec:iou-conversion}, to apply the same strictness for relative shifts that are allowed per dimension, we set this threshold to 0.374 for the evaluation of 3D bounding boxes per recording.
    Nevertheless, the corresponding IoU threshold we derived for 3D is an estimate.
    Hence, we additionally project the proposed 3D bounding boxes onto the 2D views and evaluate them per image with a threshold of 0.5 to directly compare to standard 2D detection.

    \begin{table}[t]
        \caption{\emph{Experimental results of the different networks evaluated on the test set.} For the multi-view networks the evaluation was done with the proposed 3D bounding boxes and their projections onto the 2D views.}
        \label{tbl:results}
        \centering
        \begin{tabular}{@{}l@{\hspace{1cm}}ccccc@{}}
            \toprule
            Method & Single-view & \multicolumn{2}{c}{MX-RCNN\textsubscript{avg}} & \multicolumn{2}{c}{MX-RCNN\textsubscript{max}} \\
            \cmidrule(lr){3-4}\cmidrule(lr){5-6}
            Evaluation     & 2D       & 3D       & 2D       & 3D       & 2D \\
            \midrule
            Weapon AP      & 85.56 \% & \textbf{92.28 \%} & 90.32 \% & 89.01 \% & 87.73 \% \\
            Glassbottle AP & 96.90 \% & \textbf{98.84 \%} & 95.37 \% & 98.74 \% & 95.62 \% \\
            Mean AP        & 91.23 \% & \textbf{95.56 \%} & 92.84 \% & 93.88 \% & 91.68 \% \\
            \bottomrule
        \end{tabular}
    \end{table}

    \begin{figure}[t]
        \centering
        \subfloat[\textit{Weapon} class.]{\includegraphics[height=4.2cm]{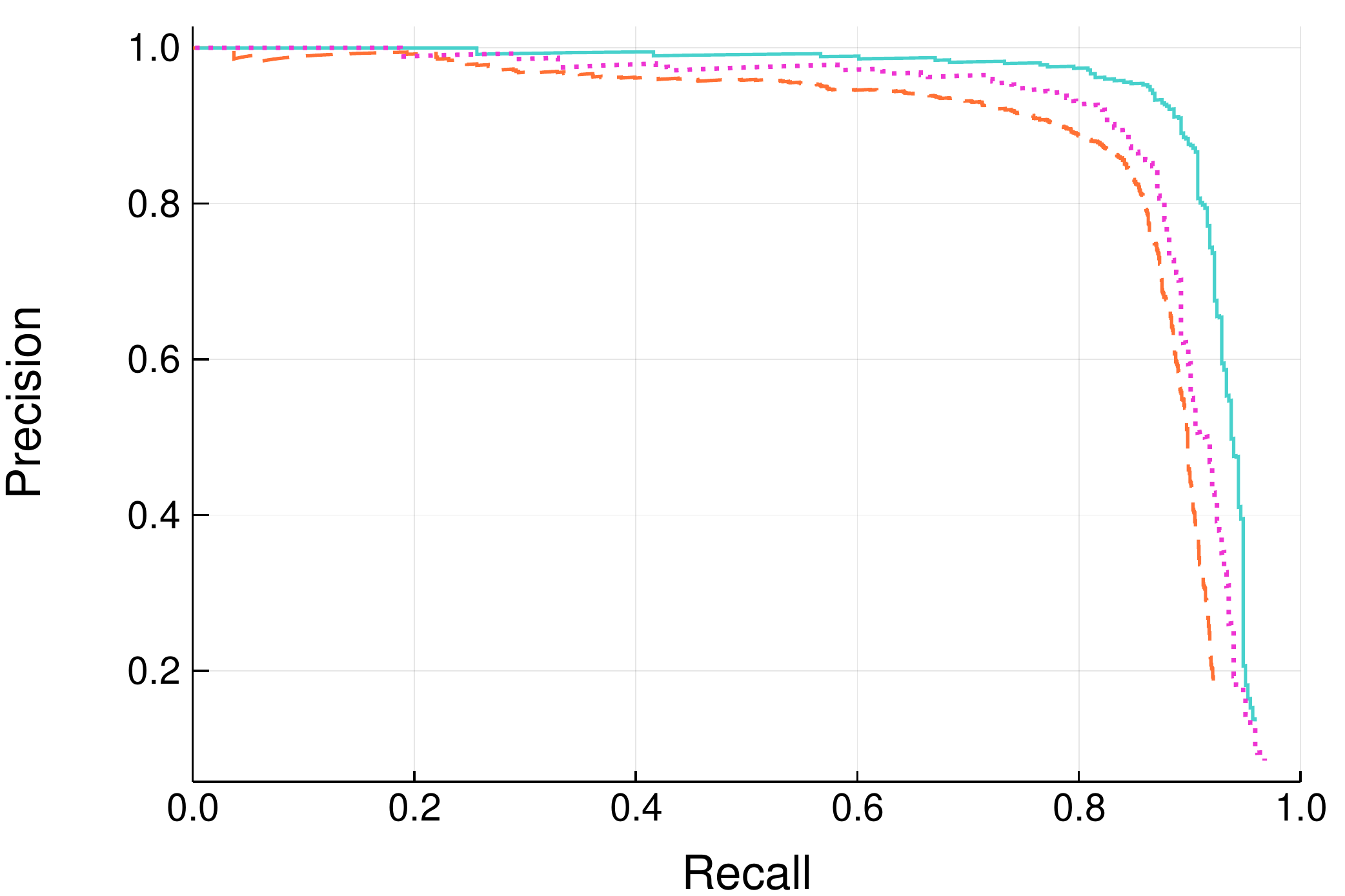}}
        \subfloat[\textit{Glassbottle} class.]{\includegraphics[height=4.2cm, trim=19mm 0 0 0, clip]{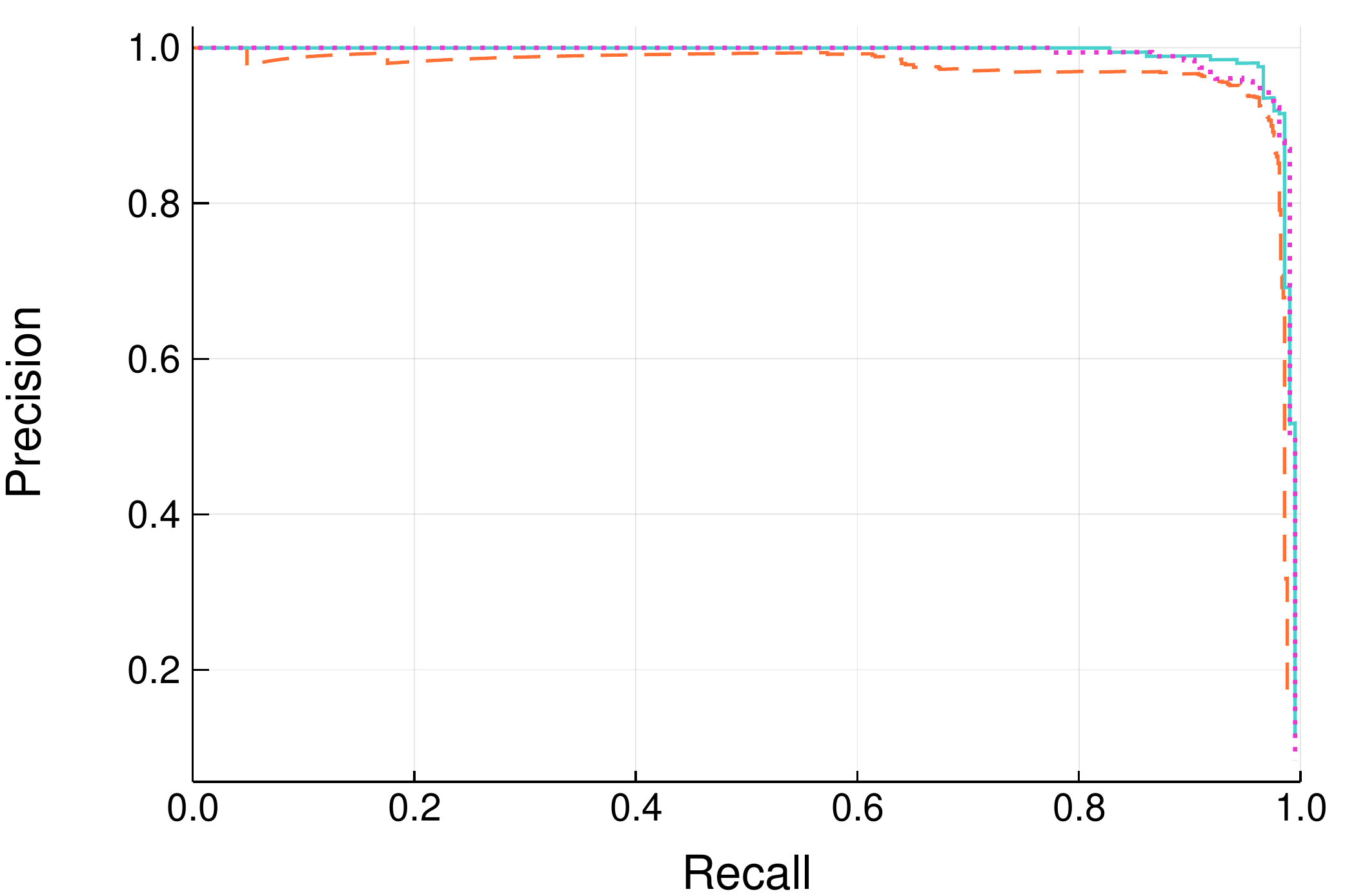}}
        \caption{\textit{Precision-recall curves of the different networks evaluated on the test set}.
        The plots show the precision-recall curves of the single-view (\textit{orange, dashed}), MX-RCNN\textsubscript{avg} (\textit{turquoise, solid}) and MX-RCNN\textsubscript{max} (\textit{pink, dotted}) networks.
        For the multi-view networks the precision-recall curves are shown for the evaluation in 3D.}
        \label{fig:precrec}
    \end{figure}

\subsection{Experimental results}
    The standard single-view Faster R-CNN \cite{Ren2017} used as a baseline reaches 91.2\% mean average precision (mAP) per image on the test set (average over classes).
    With 93.9\% mAP per recording, our MX-RCNN\textsubscript{max} is 2.7\% points better than the baseline when evaluating in 3D and 0.5\% points better when projected to 2D (91.7\% mAP per image).
    Using a weighted average in the multi-view pooling layer of MX-RCNN (MX-RCNN\textsubscript{avg}) shows consistently better detection accuracy than MX-RCNN\textsubscript{max} except for the AP of the \textit{glassbottle} class projected to 2D.
    With an mAP per recording for 3D evaluation of 95.6\%, our MX-RCNN\textsubscript{avg} is 4.4\% points better than the baseline when evaluating in 3D and 1.7\% better when projected to 2D (mAP of 92.9\%).
    Detailed numbers of the evaluation can be found in \cref{tbl:results}.
    Additionally, precision-recall curves are provided in \cref{fig:precrec} to allow studying the accuracy at different operating points.
    We observe that the ranking of the curves of the different networks within each object class is consistent with their AP values.
    Moreover, it becomes apparent that the benefit of the proposed multi-view approach is particularly pronounced in the important high-recall regime.

    \setlength{\tabcolsep}{6pt}
    \captionsetup[subfloat]{position=top}
    \begin{table}[t]
        \caption{\emph{Tolerance of MX-RCNN for disabling individual views when evaluating on the test set.} The proposed 3D bounding boxes were directly evaluated without reprojection to 2D.}
        \label{tbl:ablation}
        \vspace{-12pt}
        \centering
        \subfloat[MX-RCNN\textsubscript{avg}]{
            \begin{tabular}{@{}l@{\hspace{0.5cm}}ccccc@{}}
                \toprule
                Views          & all & w/o bottom left & w/o bottom right & w/o right side \\
                \midrule
                Weapon AP      & 92.28 \%  & 82.34 \% & 85.57 \% & 53.19 \% \\
                Glassbottle AP & 98.84 \%  & 91.58 \% & 98.59 \% & 92.42 \% \\
                Mean AP        & 95.56 \%  & 86.96 \% & 92.08 \% & 72.80 \% \\
                \bottomrule
            \end{tabular}}\\
        \subfloat[MX-RCNN\textsubscript{max}]{
            \begin{tabular}{@{}l@{\hspace{0.5cm}}ccccc@{}}
                \toprule
                Views          & all & w/o bottom left & w/o bottom right & w/o right side \\
                \midrule
                Weapon AP      & 89.01 \%  & 74.40 \% & 83.69 \% & 58.86 \% \\
                Glassbottle AP & 98.74 \%  & 89.89 \% & 97.93 \% & 77.26 \% \\
                Mean AP        & 93.88 \%  & 82.14 \% & 90.81 \% & 68.06 \% \\
                \bottomrule
            \end{tabular}}
    \end{table}

\subsection{Ablation study}
    To test the importance of the different views in our multi-view setup on the final detection result, we disabled individual views while evaluating and in the mapping provided to the multi-view pooling layer, respectively.
    The evaluation was done for both variants of the multi-view pooling layer on the test set and we directly compared the proposed 3D bounding boxes to 3D ground-truth annotations with 0.374 as IoU threshold without reprojecting them to 2D.
    The detailed results can be found in \cref{tbl:ablation}.
    We notice that the tolerance for missing views from below the tunnel is higher than for a missing side view.
    Also, the impact is more pronounced on weapons whose appearance is more affected by out-of-plane rotations.
    In general, the use of weighted averaging when combining features in the multi-view pooling seems to be more fault tolerant than the use of a weighted maximum, with the exception of weapons in combination with a missing side view.
    The results show that the network indeed relies on all views to construct the feature volume and to propose and validate detections.

	\section{Conclusion}
\label{sec:conclusion}
In this paper we have introduced MX-RCNN, a multi-view end-to-end trainable object detection pipeline for X-ray images.
MX-RCNN is a two stage detector similar to Faster R-CNN \cite{Ren2017} that extracts features from all views separately using a standard CNN backbone and then fuses these features together to shape a 3D representation of the object in space.
This fusion happens in a novel multi-view pooling layer, which combines all individual features leveraging the geometry of the X-ray imaging setup.
An experimental analysis on a dataset of carry-on luggage containing glass bottles and hand guns showed that when trained with the same annotations, MX-RCNN outperforms Faster R-CNN applied to each view separately and is computationally cheaper than separate processing of all views.
We also showed in an ablation study that the method works by far better when the view angles do not all fall in one line (degenerate 3D case), showing that the pipeline is indeed leveraging its 3D feature representation.

	{\small
	\subsubsection{Acknowledgements.}
	The authors gratefully acknowledge support by Smiths Heimann GmbH.
	}

	{\small
	\bibliographystyle{splncs04}
	\bibliography{short,papers,external,local,local2}
	}
\end{document}